# HIGH DYNAMIC RANGE IMAGE FORENSICS USING CNN

*Yongqing Huo, Xiaofeng Zhu*

School of Information and Communication Engineering, University of Electronic Science and Technology of China, Chengdu 611731 China,
hyq098132@uestc.edu.cn

## ABSTRACT

High dynamic range (HDR) imaging has recently drawn much attention in multimedia community. In this paper, we proposed a HDR image forensics method based on convolutional neural network (CNN).To our best knowledge, this is the first time to apply deep learning method on HDR image forensics. The proposed algorithm uses CNN to distinguish HDR images generated by multiple low dynamic range (LDR) images from that expanded by single LDR image using inverse tone mapping (iTM). To do this, we learn the change of statistical characteristics extracted by the proposed CNN architectures and classify two kinds of HDR images. Comparision results with some traditional statistical characteristics shows efficiency of the proposed method in HDR image source identification.

## 1. INTRODUCTION

Due to continuous development of multimedia technology, image forensics has become a popular research field. Most of the methods are designed for detecting the authenticity and integrity of 8-bit low dynamic range (LDR) images. Meanwhile, in the past decade, high dynamic range imaging has been developed rapidly in both theory and technology. It is convenient for people to get a high dynamic range (HDR) image which has wider gamut of luminance and better visual quality from various multimedia applications. However, the cost of this convenience is safety and reliability. How to detect the authenticity and integrity of the HDR images become an urgent problem.

Currently, the most popular way to generate HDR image is to combine several LDR images with different exposures from a same scene [1], [2]. By setting different weights to bright or dark region of different exposures, we can gain a HDR image which can present the details in each region of the scene. To simplify the procedure, researchers proposed the inverse tone mapping (iTM) to expand a single exposure LDR image to a HDR image directly [3],. Compared with the HDR image produced by multiple LDR images, HDR image expanded from a single LDR image has the same visual quality and was hardly distinguished by naked eyes. Image source identification, a branch of image forensics, may draw a clear distinction between the two kinds of HDR images. For the sake of clarity and brevity, in the rest of the paper we replace the two kinds of HDR image with "mHDR" and "iHDR", respectively.

Since convolutional neural network became the most popular image classification approach, researchers successfully explore to use deep learning frameworks in image forensics. Tuama et al. [4] proposed a deep learning approach for camera model identification by training a 7 layers CNN. Chen et al. [5] also used a 9 layers CNN to detect median filtered image under different JEPG compression quality factor. It's proven to have great potential in image forensics field.

In this paper, we performed mHDR image and iHDR image classification based on CNN. To do this, we collect and create a HDR image dataset. The proposed network learns and detects statistical changes of the luminance channel of HDR image automatically. Through this method, we achieve HDR image source forensics using deep learning that has not been properly considered in existing technique.

The rest of the paper is organized as follow. In section 2, we review several HDR image producing methods and existing forensics detection methods. Section 3 explains the details of our CNN architecture. We describe the performance and results in Section 4. Finally, conclusions and further discussions are presented in the last section.

## 2. BACKGROUND AND REALTED WORK

### 1.1 HDR image capturing

Because of the restriction of camera sensors' dynamic range, the complicated scenes taken by common digital camera are easy to be overexposed or underexposed. Image synthesis was introduced into capturing HDR image for solving the problem. Directly fusing the overexposed image and underexposed image can preserve the complete information of the scene. Vonikakis [6] proposed a method to fuse two or more different exposure LDR images to a HDR image using different weights which keep well-exposed pixels from each exposure. In [7] Raman et al. choose the bilateral filtered L-channel of one of the multi-exposure images to obtain the compositing weights to generate HDR images as precise as the scene human visual system (HVS) reflected.

To avoid ghost and halos in the resulting HDR image, inverse tone mapping operator (iTMO) inspired by TMO is proposed to generate HDR image from one LDR image. Akyuz et al. [8] presented a simple and global linear expansion method, which is suitable for well exposed images. In [9] and [10], the overexposed areas are handled especially using a designed function. Huo et al., [11] use a non-linear sigmoid-like function based on HVS to expand input LDR image, which is efficient for highlight region. Most of exiting iTMOs focus on enhancing the luminance in over-exposed regions with less effort on the process of the well-exposed regions, which influences HDR image quality. Therefore, the algorithms in [12] and [13] implement expanding image from both highlight and dark regions.

### 1.2 HDR image forensics

There have rarely been conducted forensic studies on HDR images. A mHDR image and four iHDR images generated from the same scene are shown in Fig. 1. From the theory of capturing method, mHDR image has more luminance details compared to iHDR image, since it is composed by multi-exposure LDR images. Conversely, the pixels of iHDR image are only expanded from pixel values between 0-255. In [14] the difference of the joint histogram of two kinds of HDR images is shown .It indicates that the joint histogram of iHDR image is sparser than that of mHDR image, especially in small pixel values. And unlike the mHDR image, there exist gaps or peaks in the joint histogram of iHDR images.

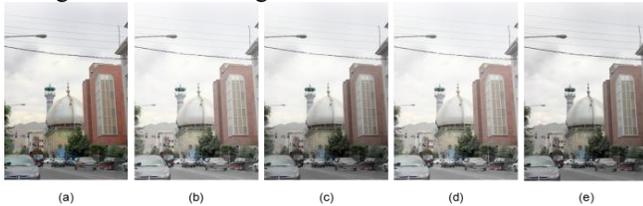

**Fig 1.** Different type HDR image generated from a same scene.(a).mhdr, (b).Akyuz's iTMO(c).Huo's iTMO (d).Kovaleski's iTMO (e).Kuo's iTMO

Therefore, Wei et al. [14] attempt to use the joint histogram characteristic to distinguish iHDR image and mHDR image. However, it has poor performance. Then they use local high-order statistics (LHS), which is applied in face recognition previously, to classify mHDR image and iHDR image. They extract the local feature to fit a Gaussian mixture model (GMM), whose parameters are taken to compute Fisher scores as the presentation of an image. Employment of SVM brings an excellent classification result. This is the only method proposed on HDR image forensics problem. However, fitting GMM and computing Fisher scores are resource-consuming and the method mentioned above does not have enough robustness. Outperforming traditional approaches and applying on various LDR image forensics problems, CNN may be more effective in this situation.

## 3. PROPSED METHOD

### 3.1 Pre-processing

When manipulating images in various ways, there are more or less signs left by tampers. By detecting these signs, such as inconsistency in image region, JPEG compressing coefficients and histograms, researchers can distinguish the tampered images. Different from LDR image forensics, discriminating mHDR image from iHDR image is usually conducted on luminance domain since the manipulation of generating HDR images mainly function on luminance channel. Though the difference of luminance can be shown on RGB channel indirectly, the objects and edges of the scene can be interference when using CNNs to learn features automatically. Without loss of generality, we only consider the luminance component of HDR images. The luminance channel can be computed as follow:

$$L = 0.2126*R + 0.7152*G + 0.0722*B$$

HDR image pixel values are not integers in range of 0-255 like LDR image but float point numbers. The maximum and minimum luminance values of different HDR image vary largely across different scenes. In our dataset, the dynamic range of different image varies from 3 to 7 orders of magnitude. To compensate for the image-dependent peak brightness in HDR image, we perform following transform:

$$l_{i,j} = \log(L_{i,j} + \varepsilon)$$

We set $\varepsilon = 10^{-6}$ to avoid singularity when taking logarithm. This log domain luminance ensures that the input of CNN is limited to a far smaller range than using luminance directly.

## 3.2 CNN architecture

Fig 2. illustrates the first CNN architecture of the proposed method. Our architecture is inspired by VGG network [15].When an image enters the proposed network, it goes through two convolutional layers and a pooling layer for three times. Followed by three fully-connected layers, the image is converted to a vector that represents the likelihood of the kind of HDR image the input belongs to.

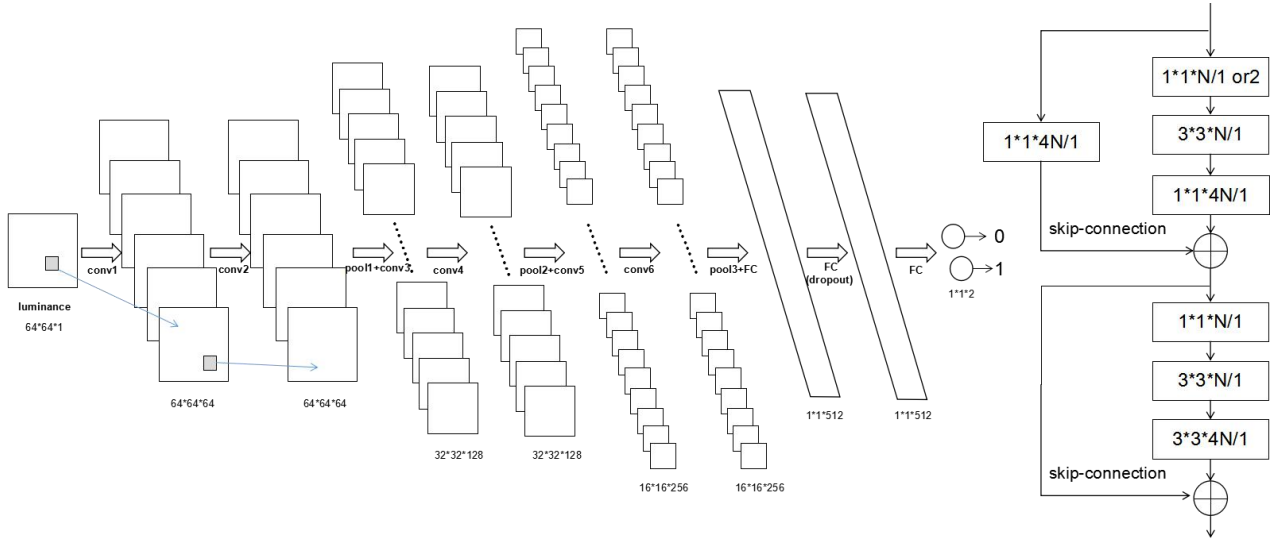

**Fig 2.** The first architecture of CNN we proposed(plain) is in the left. The residual block we used is in the right. Behind "/" is the stride of that layer. Every layer represents a convolution layer with ReLU activation function followed by batch norm layer.

Every convolutional layer in our network convolves the previous layer's output with some kernels of size 3*3 and the stride of the kernel is 1. The first two convolutional layers have 64 kernels, the second two and the third two have 128,256 kernels respectively. Each convolutional layer is followed by a non-linear activation function ReLU. After convolution, the feature maps are huge to compute and prone to over-fitting. To reduce the feature dimension, we choose max-pooling operator with the window size of 2*2 and step size of 2. The end of our CNN are three full connected layers. The first two fully connected layers have 512 neurons and the last one only has 2 neurons because our classification task has two classes. As well as dropout layer is introduced between each fully connected layer to prevent over-fitting. After the last fully connected layer, a cross entropy loss function with softmax activation is used for classification. In this way, the classification result can be fed back to guide the feature extraction automatically.

For the purpose of improvement of accuracy, we also attempt another architecture which use residual block [16] to take place of common convolution layer. Residual block is well-known by its ability of preventing network degradation when the network goes deeper. The shortcut-connection ensures the network learns identity mapping of the input and the output of residual block. The residual block we used is also shown in Fig 2.

The second network we proposed is similar to the first one. We replace all convolution layers in the first network with residual blocks. Before all residual blocks, we add a large kernel convolution layer and a overlapping average pooling layer before residual blocks. Instead of the first two fully-connected layers, we use a global average pooling layer to reduce the dimension of the feature maps which used to classify.

## 4. EXPERIMENTAL RESULTS

### 4.1 Experiment parameters and settings

For training the proposed networks, we collect some mHDR images and generate some iHDR images to create a dataset. We collected 458 high-resolution mHDR images from the follow 4 sources [15]:
- Meylan created 14 mHDR images;
- Fairchild created 106 mHDR images;
- HDRSID dataset contains 232 mHDR images;
- Online data includes 106 images.

Among the above mentioned images, we pick out 406 mHDR images whose scene is not very similar to each other. However, we have no the original LDR image sequences that generate mHDR images above mentioned. As a supplement, we randomly pick 406 LDR images from the image dataset "mitadobe5k [17]", which contain 5000 high-quality and high-resolution images stored in RAW format.We apply four kinds of iTMOs [9,10,11,12] to 406 LDR images to create iHDR images.

Although creating a dataset including 812 HDR images, it is nowhere near enough to train CNN. Moreover, our HDR images size varies from 1024*1280 to 5202*3465, which affects the depth of CNN at some extent. We firstly resize these HDR images according to their original size. Then, two kind of HDR images are divided into 64*64 sub-images blocks and shuffled completely before learning. From each kind of HDR image, we pick out 40 images for verification of network before dividing and we ensure every type of iHDR image or mHDR image is in the verification dataset 1.0. 60000 image blocks is used to train the network and the ratio of mHDR blocks and iHDR blocks is 1:1. The remaining sub-image blocks constitute verification dataset 2.0. 50 epochs are performed for training. In the verification processing, the majority voting system (MVS) is applied to the sub-images to make a judgement on the whole image in verification 1.0.

### 4.2 Experiment evaluation

We first discuss the advantage of using log domain luminance as the input of the network. Table 1 shows the accuracy of using normalization pixel values or using log domain luminance of the same image block respectively. We notice that the pixel value leads to an unstable outcome in the early training phase. When the network reaches convergence, the network takes pixel value as input has poorer performance than that takes luminance as input. The ROC curves of two architectures are shown in Fig 3. However, we found that the common convolution layer in the first architecture seems more useful than the first residual block. It is because the feature map size of Res1 decreases faster than that of Plain, which leads to that the network cannot fully study larger feature.

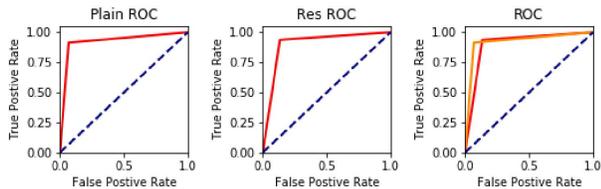

**Fig 3.** ROC curves of our proposed two network

After checking the misjudged image block of the two kinds of HDR image, we notice that the error usually appears in some smooth regions or some sections that have several strong edges no matter in light source region or reflection region. And these regions almost are the brightest or darkest part of the whole image as we mentioned above about the joint histogram. Moreover, It is no surprise that the detection accuracy of mHDR image is slightly higher than that of iHDR image since the kind of mHDR image is larger than the kind of iHDR image.

Due to lack of classical state of art method except for the method proposed by Wei in [13], we extract some statistical features which are usually used in image classification to be comparison. Histogram of oriented gradient (HOG) is one of the most useful descriptors that used in object detection. Local binary pattern (LBP) is famous of detecting texture of the image. Steganalytic feature SPAM is also possible to fulfill this forensic task.

**Table 1.** The accuracy of the same network architecture with different input. Plain represents the first network use convolution layer as base block. Res1 use residual block as the base unit of the network respectively.

|  | Plain | Res |
|---|---|---|
| Normalized pixel value | 92.55% | 90.71% |
| Log domain luminance | 94.15% | 93.36% |

**Table 2.** Performance on verification dataset.In verification dataset 1.0, the accuracy after MVS is in the bracket.

| | | | Plain | Res |
|---|---|---|---|---|
| Verification 1.0 | ihdr | acc | 93.26(100) | 86.61(100) |
| | | AUC | 92.29 | 90.16 |
| | mhdr | acc | 91.33(100) | 93.70(100) |
| | | AUC | 93.17 | 94.55 |
| Verification 2.0 | accuracy | | 94.09 | 93.70 |

**Table 3.** Detection performance of different methods on verification dataset 1.0

|  | CNN | HOG | LBP | SPAM | LHS[13] |
|---|---|---|---|---|---|
| accuracy | 92.75 | 60.28 | 63.65 | 69.53 | 80.26 |
| AUC | 92.63 | 64.49 | 68.86 | 77.01 | 85.33 |

Different from deep learning method, we use part of our training data to extract the three features mentioned above. For HOG, we set 16*16 pixels as a cell and 2*2 cells as a block. A 324-demensions HOG descriptor is used to represent an image. When extracting LBP, we use uniform pattern to gain a 944-demensions feature. All the features are used to train SVM classifier and we conduct grid search to find optimal parameters. The result of testing verification dataset 1.0 is shown in Table 3. The ROC curves of three methods are displayed in Fig 4.

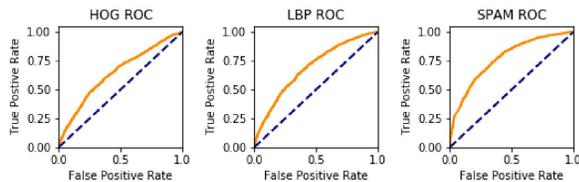

**Fig 4.** Some ROC curves of our compared statistical characteristics.

It is shown that common statistical feature does not solve the forensics problem and the proposed method achieves high detection rate and outperforms the tradition machine learning method.

## 5. CONCLUSION

In this paper, we introduced a deep neural network based forensic method which able to recognize the source of HDR image. Using features learned automatically from deep learning model, we can achieve better detection accuracy compared with traditional method. The proposed technique solved HDR image source problem by CNN for the first time, but many issues still need be considered in following work. We only choose four kinds of iHDR images to train the network which means the performance of other type of iHDR images probably not good as that we used. It is still meaningful to analyze more complicated HDR image forensics problem using CNN architecture in the future.